\documentclass[journal=jacsat,manuscript=article]{achemso}

\usepackage[version=3]{mhchem} 

\usepackage{color}
\usepackage{ulem}
\usepackage{amsfonts}
\usepackage{array}
\usepackage{booktabs}
\usepackage{multirow}
\usepackage{graphicx}

\usepackage{algorithm}
\usepackage{algpseudocode}

\abbreviations{IR,NMR,UV}

\author{Dedi Wang}
\affiliation{Biophysics Program and Institute for Physical Science and Technology, University of Maryland, College Park 20742, United States}
 
\author{Yihang Wang}
\altaffiliation{These authors contributed equally.}
\affiliation{Biophysics Program and Institute for Physical Science and Technology, University of Maryland, College Park 20742, United States}

\author{Luke Evans}
\altaffiliation{These authors contributed equally.}
\affiliation{Department of Mathematics, University of Maryland, College Park, MD 20742, USA}

\author{Pratyush Tiwary}
\email{ptiwary@umd.edu}
\affiliation{Department of Chemistry and Biochemistry and Institute for Physical Science and Technology, University of Maryland, College Park 20742, United States}
\title{From latent dynamics to meaningful representations}

\begin{document}

\begin{abstract}
  While representation learning has been central to the rise of machine learning and artificial intelligence, a key problem remains in making the learnt representations meaningful. For this the typical approach is to regularize the learned representation through prior probability distributions. However such priors are usually unavailable or are \textit{ad hoc}. To deal with this, recent efforts have shifted towards leveraging the insights from physical principles to guide the learning process. In this spirit, we propose a purely dynamics-constrained representation learning framework. Instead of relying on predefined probabilities, we restrict the latent representation to follow overdamped Langevin dynamics with a learnable transition density — a prior driven by statistical mechanics. We show this is a more natural constraint for representation learning in stochastic dynamical systems, with the crucial ability to uniquely identify the ground truth representation. We validate our framework for different systems including a real-world fluorescent DNA movie dataset. We show that our algorithm can uniquely identify orthogonal, isometric and meaningful latent representations.
\end{abstract}

\section{Introduction}
\label{sec:introduction}

The ability to learn meaningful and useful representations of data is a key challenge towards applying machine learning (ML) and Artificial Intelligence (AI) to various real world problems.  On one hand, a useful representation extracts and organizes the discriminative information from data to support effective machine learning for downstream tasks \cite{Bengio2013, Scholkopf2012, Peters2017, Kalinin2020, Ziatdinov2021}. In the realm of chemistry, such representations offer promising avenues for enhancing predictions of molecular properties\cite{Fang2022}, forecasting reactions\cite{Coley2019}, predicting pharmacological activities\cite{Sakai2021}, facilitating exploration of vast chemical spaces\cite{Kaufman2023}, and accelerating computational simulations\cite{Wang2020,Mehdi2024}. On the other hand, a meaningful representation is often more interpretable, which is fundamental to help humans trust AI. Recently, a variety of representation learning methods have been proposed based on the idea of autoencoding -- learning a mapping from high dimensional data to a low dimensional representation or latent space which is able to approximately reconstruct the original data\cite{review_AE}. However, while the ability to reconstruct the original data might make a representation \textit{useful}, it does not necessarily make it \textit{meaningful}\cite{Bengio2013}. 

Variational Autoencoders (VAEs) approach the task of learning representations for high-dimensional distributions as a form of variational inference, simultaneously enabling representation learning and sample generation\cite{VAE}. VAEs have rapidly gained popularity and become a favored framework among diverse representation learning methods\cite{review_AE, review_disentanglement, Kingma2019}.

Recently, there is growing interest in including a physics-based prior into neural networks. Refs. \citenum{Greydanus2019, Lutter2019, Cranmer2020} introduce concepts from Lagrangian or Hamiltonian mechanics into neural networks, allowing them to learn and respect conservation laws for deterministic dynamical systems. These dynamical priors have subsequently been incorporated into VAEs\cite{Toth2019, Saemundsson2020, Yang2022, Khan2022}. Incorporating domain knowledge from physics also encompasses alternative approaches, such as the use of physical laws as regularization terms to augment the loss function\cite{Raissi2019, Zhu2019, Kaltenbach2021}.

Some VAE extensions for sequential data have been proposed to incorporate temporal dependencies in sequences using recurrent neural networks or state-space models\cite{Chung2015, Hsu2017, Li2018, Girin2020}. In line with this, Ref. \citenum{Hasan2021} extends the VAE framework by incorporating a latent space shaped by a stochastic differential equation (SDE) with an isotropic diffusion coefficient. They theoretically establish that such a dynamical constraint enables the attainment of a unique representation, preserving the essence of the original data up to an isometry.

In addition to autoencoder-based methods, there exists an extensive body of literature in the field of nonlinear Independent Component Analysis (ICA) that leverages the temporal structure of data\cite{Hyvarinen2016, Hyvarinen2017, Halva2020, Morioka2021}. Recent research also proposes a unifying perspective that bridges the gap between VAEs and nonlinear ICA\cite{Khemakhem2020}.

In this work, we revisit the importance of introducing physics-based dynamics laws into representation learning. We find that attributes normally associated with meaningful representations naturally and directly arise from our physics-based framework, which we call Dynamics Constrained Auto-encoder (DynAE). To our knowledge, this is the first representation learning algorithm which regularizes the latent space purely based on its dynamical properties. It achieves latent space regularization solely through the inherent dynamics, setting it apart from existing dynamical VAE methods\cite{Hasan2021,Girin2020} that maximize the likelihood of probability distributions and still demand a prior knowledge of the latent variable's probability distribution, often assuming Gaussian distributions. This innovative approach eliminates the need for any prior information about the distribution of the latent space, a requirement that is typically unmet in complex systems. Furthermore, we tackle the inherent challenge\cite{Chen2016, WAE} posed by the VAE objective by extending the concept of the sliced Wasserstein auto-encoder. Rather than focusing on regularizing the latent variable distribution, we shift our attention to regularizing the transition density distribution. This seemingly simple adjustment unlocks the ability to learn distinctive representations for a wide array of systems characterized by diverse probability distributions, as demonstrated in our work.

Here we focus specifically on introducing priors corresponding to overdamped Langevin dynamics, deeply driven by the principles of statistical mechanics. Nevertheless, we anticipate the framework should be easily generalized to other physics-based dynamics models. For the widely adopted model of Brownian or overdamped Langevin dynamics \cite{Chandrasekhar1949}, we provide both theoretical and numerical evidence that such a dynamical constraint makes the proposed model able to recover the ground-truth latent variables up to an isometry. We validate our algorithm through a variety of stochastic models and datasets, including a fluorescent DNA movie dataset. Across these diverse examples, our results provide clear evidence that DynAE consistently outperforms competing approaches in identifying the true latent factors.

\begin{figure}[t!]
    \centering
    \includegraphics[width=0.45\textwidth]{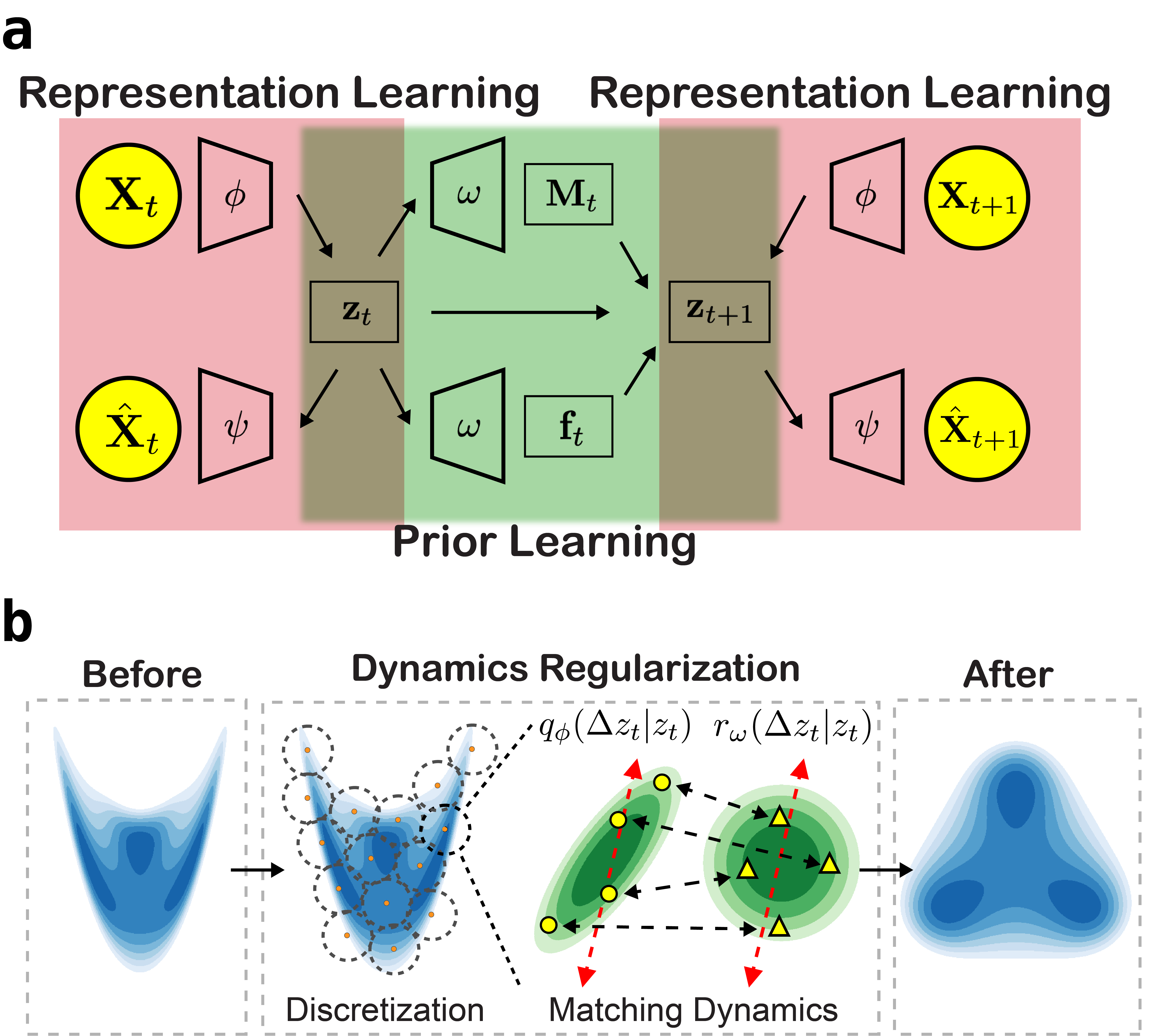}
    \caption{\textbf{a}. Network architecture used for dynamics constrained framework. The encoder $\phi$, the decoder $\psi$, the diffusion $\mathbf{M}_\omega$ and the force field $\mathbf{f}_\omega$ are nonlinear deep neural networks. \textbf{b}. A flowchart illustrating the dynamics constrained framework. To regularize latent dynamics, the latent representation is discretized into bins represented by dashed circles. The samples, depicted as yellow circles, are drawn from each bin based on the encoded transition density $q_\phi(\Delta\mathbf{z}|\mathbf{z}_t)$ and then matched with samples, shown as yellow triangles, drawn from some specific prior transition density $r_\omega(\Delta\mathbf{z}|\mathbf{z}_t)$. The randomly chosen slicing direction is indicated by the red dashed double-headed arrow. See details in the Methods section.}
    \label{fig:alg}
\end{figure}

\section{Results}
\label{sec:results}

\subsection{Incorporating the dynamics into representation learning.}\label{sec:rep_learning} 
The goal of autoencoder-based representation learning algorithms in general is to learn a low-dimensional latent representation $\mathbf{z}$ such that the original high-dimensional observations can be reconstructed from the low-dimensional representation. Most of the recent advances in representation learning are built upon the pioneering work of the VAE\cite{review_AE, VAE}. These algorithms consist of at least three important components: an encoder $q_{\phi}(\mathbf{z}|\mathbf{X})$, a decoder $q_{\psi}(\mathbf{X}|\mathbf{z})$, and a prior distribution $r_\omega(\mathbf{z})$ for the latent representation space. The encoder maps the input $\mathbf{X}$ to the representation space $\mathbf{z}$, while the decoder reconstructs the input data $\mathbf{X}$ from the representation. Both the encoder $q_{\phi}(\mathbf{z}|\mathbf{X})$ and decoder $q_{\psi}(\mathbf{X}|\mathbf{z})$ are usually parameterized by neural networks with the model parameters $\phi$ and $\psi$. In addition, the encoder is encouraged to satisfy some structure specified by the prior distribution $r_\omega(\mathbf{z})$ on the latent space, where $\omega$ denotes trainable parameters for the prior. The objective function $ \mathcal{L}_{rep}$ of VAE, where the subscript \textit{rep} signifies the representation learning process, can be written as follows:
\begin{equation} 
\begin{aligned}
\label{eq:obj}
\mathcal{L}_{rep}(\phi,\psi;\omega)&\equiv\\ \mathbb{E}_{\hat{p}(\mathbf{X})}\mathbb{E}_{q_{\phi}(\mathbf{z}|\mathbf{X})}&\bigg[-\log q_{\psi}(\mathbf{X}|\mathbf{z}) 
+ \beta\log\frac{q_{\phi}(\mathbf{z}|\mathbf{X})}{r_{\omega}(\mathbf{z})}\bigg]\\
&=\mathcal{L}_{REC}+\beta\mathcal{L}_{REG},
\end{aligned}
\end{equation}
where $\hat{p}(\mathbf{X})\equiv\frac{1}{N}\sum_{n=1}^N\delta_{\mathbf{X}_n}$ is the empirical data distribution, and $N$ is the number of data points. This objective function can be divided into two parts: The first term $\mathcal{L}_{REC}$ measures the ability of our representation to reconstruct the input, while the second term $\mathcal{L}_{REG}$ can be interpreted as the complexity penalty that acts as a regulariser. Such a trade-off between the prediction capacity and model complexity can then be controlled by a hyper-parameter $\beta\in[0,\infty)$. Compared to the original VAE objective where $\beta=1$, a modified version named $\beta$-VAE shows a better performance in terms of disentanglement with an appropriately tuned $\beta$\cite{beta-VAE}.

Minimizing the objective in Eq. \ref{eq:obj} requires a definition of the prior distribution $r_{\omega}(\mathbf{z})$ which can be difficult. Choosing a simplistic prior like the standard Gaussian distribution commonly used in VAEs could lead to over-regularization. resulting in poor latent representations. Conversely, adopting an overly flexible prior, such as a mixture of Gaussians prior \cite{Dilokthanakul2016}, VampPrior \cite{VampPrior}, or normalizing flows \cite{NF_review}, might defeat its purpose and no longer serve as a regularizer for the learned latent space.

To deal with this dilemma, a compelling approach is to integrate physics-based priors into the representation learning process. Instead of using a predefined prior distribution to regularize the latent space, we can enforce the latent representation to follow a specific yet generic class of dynamics. Our inspiration here stems from physics, where even though different systems can and likely will have different probability distribution along the same low-dimensional projection $z$, they will obey the same dynamics equations, such as Newton's second law in classical mechanics and Schrödinger's equation in quantum mechanics. Therefore, we believe this is a more natural and generic constraint for representation learning. 

Indeed, incorporating a dynamical model into VAEs is not entirely a novel concept. In Ref. \citenum{Hasan2021}, the authors extended the VAE framework to accommodate pairwise consecutive time series observations denoted as $(\mathbf{X}_{t},\mathbf{X}_{t+\Delta t})$ and their associated latent variables $(\mathbf{z}_{t},\mathbf{z}_{t+\Delta t})$, thereby accounting for dynamic aspects. Throughout this paper, we refer to this approach as SDE-VAE. The objective function $\mathcal{L}_{rep}$ of this generalized VAE can be written as follows:
\begin{equation} 
\begin{aligned}
\label{eq:SDE-VAE obj}
\mathcal{L}_{rep}(\phi,\psi;\omega)&\equiv\\ \mathbb{E}_{\hat{p}(\mathbf{X}_t,\mathbf{X}_{t+\Delta t})}\mathbb{E}_{q_{\phi}(\mathbf{z}_{t},\mathbf{z}_{t+\Delta t}|\mathbf{X}_{t},\mathbf{X}_{t+\Delta t})}&\bigg[-\log q_{\psi}(\mathbf{X}_{t},\mathbf{X}_{t+\Delta t}|\mathbf{z}_t,\mathbf{z}_{t+\Delta t})\\ 
&+\beta\log\frac{q_{\phi}(\mathbf{z}_t,\mathbf{z}_{t+\Delta t}|\mathbf{X}_t,\mathbf{X}_{t+\Delta t})}{r_{\omega}(\mathbf{z}_{t},\mathbf{z}_{t+\Delta t})}\bigg].
\end{aligned}
\end{equation}
A number of assumptions are made to further simplify the objective and enable its efficient training (see detailed discussion in Supplementary Methods in the Supporting Information). This objective function (Eq. \ref{eq:SDE-VAE obj}) can also be viewed as a special case of the objective functions used in other dynamical VAE models, which allow for higher order of temporal models\cite{Girin2020}.

While these existing methods hold promise, they still grapple with the inherent limitations of VAE approaches, and thus fail to attain meaningful representations in practice. First, even though these methods introduce a dynamical model into the algorithm, they maximize the likelihood of probability distributions, which again necessitates knowledge of the prior probability distribution of the latent variables $r_{\omega}(\mathbf{z})$. Our current work removes such a requirement. In practice, determining the exact form of this probability distribution, especially for complex systems, remains elusive \textit{a priori}.
Assuming the prior $r_{\omega}(\mathbf{z})$ to be Gaussian, as seen in Ref. \citenum{Hasan2021} and similar works, could result in over-regularization, leading to poor representations\cite{Hoffman2016}.
Moreover, the VAE-based objective in Eq. \ref{eq:SDE-VAE obj} itself poses difficulties in learning useful representation and achieving effective regularization. On one hand, such regularization can yield solutions that disconnect the latent space from input data, potentially resulting in issues like posterior collapse when coupled with potent decoders\cite{Chen2016, WAE}. On the other hand, the use of the Kullback–Leibler (KL) divergence between the prior distribution and the posterior distribution in practice does not inherently ensure that the latent space adheres to the prior distribution or the specified dynamics. In fact, the learned aggregated posterior distribution frequently diverges from the assumed latent prior\cite{Hoffman2016,Rosca2018}.

To address the limitations inherent in existing methods, we present a representation learning framework summarized in Fig. \ref{fig:alg} that relies exclusively on intrinsic dynamical properties. This innovative approach obviates the necessity for any prior knowledge regarding the distribution of the latent space, a requirement that is frequently unfulfilled in complex systems. Additionally, we tackle the intrinsic challenge presented by the VAE objective by extending the concept of the sliced Wasserstein auto-encoder. This allows us to move beyond the confines of the traditional maximum likelihood framework employed in VAE-based representation learning.

\subsection{A representation learning framework exclusively driven by dynamics.}
For Markovian processes, their dynamics can be described by their transition density $p(\Delta\mathbf{z}|\mathbf{z}_t)$ with $\Delta \mathbf{z} \equiv \mathbf{z}_{t+\Delta t} - \mathbf{z}_t$. For deterministic dynamical processes, the transition density would equal a delta function. A tempting idea then is to match the encoded transition density $q_\phi(\Delta\mathbf{z}|\mathbf{z}_t)$ to some specific prior transition density $r_\omega(\Delta\mathbf{z}|\mathbf{z}_t)$. This can be done by minimizing the sliced-Wasserstein distance between these two conditional probability distributions\cite{WAE, SWAE}. Compared to the commonly used KL divergence, the Wasserstein distance offers greater stability and consistently yields finite and intuitive values, even in cases involving non-overlapping distributions where KL divergence falls short. Moreover, it possesses fundamental properties, including symmetry and the triangle inequality, making it a true metric. The sliced-Wasserstein distance shares these advantageous properties with the Wasserstein distance, while being much simpler to compute\cite{Kolouri2019}. With this we propose a dynamics constrained regularizer $\mathcal{L}_{REG}=\mathcal{D}_{SW}(q_\phi(\Delta\mathbf{z}_t|\mathbf{z}_t),r_\omega(\Delta\mathbf{z}_t|\mathbf{z}_t))$. It's worth emphasizing that this regularizer relies solely on the latent dynamics, rendering it independent of any prior knowledge regarding the latent space distribution. In practice, the sliced-Wasserstein distance $\mathcal{D}_{SW}(q_\phi(\Delta\mathbf{z}_t|\mathbf{z}_t),r_\omega(\Delta\mathbf{z}_t|\mathbf{z}_t))$ is approximated by discretizing the latent space into bins. In the Methods section below, we give a full discussion of the proposed dynamics constrained regularizer.

For simplicity, in this work we consider only the case of mappings where the latent variable $\mathbf{z}$ is a deterministic function of the input data $\mathbf{X}$ given as $\mathbf{z}=\phi(\mathbf{X})$, but the algorithm should be easily generalized to stochastic encoders. Let $\psi$ denote the deterministic decoder that maps the latent variable $z$ back to the original data space. In this way, the reconstruction loss $\mathcal{L}_{REC}$ can be replaced by the $L_2$-loss. We can obtain the objective function $\mathcal{L}_{rep}$ for learning dynamics constrained latent representations:
\begin{equation} 
\begin{aligned}
\label{eq:dynamics_rep_obj}
\mathcal{L}_{rep}=\mathbb{E}_{\hat{p}(\mathbf{X}_t,\mathbf{X}_{t+\Delta t})}\bigg[||\psi(\phi(\mathbf{X}_t))-\mathbf{X}_t||_2^2\\
+||\psi(\phi(\mathbf{X}_{t+\Delta t}))-\mathbf{X}_{t+\Delta t}||_2^2\bigg]\\
+\beta \mathcal{D}_{SW}(q_\phi(\Delta\mathbf{z}_t|\mathbf{z}_t),r_\omega(\Delta\mathbf{z}_t|\mathbf{z}_t)),
\end{aligned}
\end{equation}
where we can tune the hyper-parameter $\beta$ to force the latent space to follow specific dynamics given the prior transition density $r_\omega(\Delta\mathbf{z}_t|\mathbf{z}_t)$. In the following, we will discuss how to choose such a prior transition density. 

\subsection{Learning the dynamics prior from samples.}
\label{sec:Langevin_prior_learning}
We focus in this paper on stochastic systems governed by the overdamped Langevin equation or Brownian dynamics as the underlying dynamics, but the framework should be more generally applicable to other dynamical systems. Consider a system where the representation $\mathbf{z}$ obeys Brownian dynamics,
\begin{equation} 
\begin{aligned}
\label{eq:brownian_motion}
d\mathbf{z}_t=&\left[\mathbf{M}(\mathbf{z}_t)\mathbf{f}(\mathbf{z}_t)+\nabla\cdot \mathbf{M}(\mathbf{z}_t)\right]dt\\
&+\sqrt{2\mathbf{M}(\mathbf{z}_t)}d\mathbf{w}_t,
\end{aligned}
\end{equation}
where $\mathbf{f}(\mathbf{z})$ is the force field experienced along $\mathbf{z}$, $\mathbf{M}(\mathbf{z})$ is the diffusion matrix, and $d\mathbf{w}$ is a Brownian motion in $\mathbb{R}^d$. Further, the force field $\mathbf{f}(\mathbf{z})$ can also be given by $\mathbf{f}(\mathbf{z}) = -\nabla F(\mathbf{z})$ where we have introduced the free energy $F(\mathbf{z})=-kT\log r_\omega(\mathbf{z})$. If we assume the diffusion matrix in the learned representation space is diagonal, Eq. \ref{eq:brownian_motion} can be simplified as:
\begin{equation} 
\begin{aligned}
\label{eq:simplified_brownian_motion}
d z_i=\left[M_{ii}(\mathbf{z}_t)f_i(\mathbf{z}_t)+\frac{\partial M_{ii}(\mathbf{z}_t)}{\partial z_i}\right]dt+\sqrt{2M_{ii}(\mathbf{z}_t)dt} \epsilon_i,
\end{aligned}
\end{equation}
where $\epsilon_i\sim \mathcal{N}(0,1)$. This assumption of diagonal diffusion matrix allows us to obtain a set of "orthogonal" latent variables. For simplicity, in further derivations we absorb the time unit into the diffusion matrix $\mathbf{M}$ and set $dt=\Delta t=1$. The prior transition density of the system $r_\omega(\Delta\mathbf{z}_t|\mathbf{z}_{t})$ parameterized by $\omega$ can then be obtained:
\begin{equation} 
\begin{aligned}
\label{eq:transition}
\log r_\omega(\Delta\mathbf{z}_t&\equiv \mathbf{z}_{t+1}-\mathbf{z}_{t}|\mathbf{z}_{t})=-\frac{1}{2}\sum_{i}\Bigg[\log M_{ii}(\mathbf{z}_{t})\\
&+\frac{\left(\Delta z_i-M_{ii}(\mathbf{z}_t)f_i(\mathbf{z}_t)-\frac{\partial M_{ii}(\mathbf{z}_t)}{\partial z_i}\right)^2}{2M_{ii}(\mathbf{z}_{t})}\Bigg],
\end{aligned}
\end{equation}
where both the force field $\mathbf{f}(\mathbf{z})=\mathbf{f}_\omega(\mathbf{z})$ and the diffusion matrix $\mathbf{M}=\mathbf{M}_\omega(\mathbf{z})$ are nonlinear functions of the representation $\mathbf{z}$ parameterized by neural network parameters $\omega$. Given the pairwise samples $\{\mathbf{z}_t,\mathbf{z}_{t+\Delta t}\}$, both the force field $\mathbf{f}_\omega(\mathbf{z})$ and the diffusion matrix $\mathbf{M}_\omega(\mathbf{z})$ can be inferred through likelihood maximization or equivalently by minimizing the following prior loss:
\begin{equation} 
\begin{aligned}
\label{eq:stochastic_prior_obj}
\mathcal{L}_{prior}(\omega;\phi,\psi)=-\mathbb{E}_{\hat{p}(\mathbf{z}_t,\mathbf{z}_{t+\Delta t})} \log r_{\omega}(\Delta\mathbf{z}_t|\mathbf{z}_t).
\end{aligned}
\end{equation}

Once the estimated force field $\mathbf{f}_\omega(\mathbf{z})$ and the diffusion matrix $\mathbf{M}_\omega(\mathbf{z})$ are obtained, we can easily draw samples $\{\Delta\tilde{\mathbf{z}}_t\}$ from the prior transition density $r_\omega(\Delta\mathbf{z}_t|\mathbf{z}_{t})$ using Eq. \ref{eq:simplified_brownian_motion}, which can then be used to regularize the representation learnt through Eq. \ref{eq:dynamics_rep_obj}.

In fact, given this generalizable framework, we can introduce additional constraints into the prior. For instance, in our experiments we observed that manually enforcing a constant diffusion $M_{ij}^* \equiv \delta_{ij}$ when optimizing Eq. \ref{eq:dynamics_rep_obj} largely stabilized the optimization process and enabled the algorithm to focus more effectively on the underlying structure of the data. Namely, we incorporate an inductive bias into the learning process and regularize the latent representation by directly using the samples $\{\Delta\tilde{\mathbf{z}}_t\}$ generated from the simplified Langevin dynamics obtained after using $M_{ij}^* \equiv \delta_{ij}$ in Eq. \ref{eq:simplified_brownian_motion}:
\begin{equation} 
\begin{aligned}
\label{eq:constant_diffusion_langevin_flow}
\Delta\tilde{{z}}_i=f_i(\mathbf{z})+\epsilon_i.
\end{aligned}
\end{equation}
By learning a latent space with isotropic and homogeneous diffusion as in Eq. \ref{eq:constant_diffusion_langevin_flow},  we can aim to preserve some underlying geometric properties of the data as we show later through numerical examples.

\begin{table}[th]
\centering
\caption{Summary of the models used in numerical experiments with detailed descriptions in the Supplementary Methods of the Supporting Information. A brief discussion of other relevant works is available in the Supplementary Discussion of the Supporting Information}.
\label{tab:models}
\begin{tabular}{>{\centering\arraybackslash}p{1.8cm} >{\centering\arraybackslash}p{2.6cm} >{\centering\arraybackslash}p{2.6cm} >{\centering\arraybackslash}p{3.8cm} >{\centering\arraybackslash}p{3.6cm}}
\toprule
 Model & $\beta$-VAE\cite{beta-VAE} & SWAE\cite{SWAE} & SDE-VAE\cite{Hasan2021} & DynAE\\ 
\midrule
 Mapping & Stochastic & Deterministic & Stochastic & Deterministic\\
 Dynamics & No & No & Yes & Yes\\
 \multirow{2}{*}{$\mathcal{L}_{REG}$} & \scalebox{0.8}{\multirow{2}{*}{$\mathcal{D}_{KL}(q_\phi(\mathbf{z}|\mathbf{X}),r_\omega(\mathbf{z}))$}} & \scalebox{0.8}{\multirow{2}{*}{$\mathcal{D}_{SW}(q_\phi(\mathbf{z}),r_\omega(\mathbf{z}))$}} & \scalebox{0.8}{$\mathcal{D}_{KL}(q_\phi(\mathbf{z}_t,\mathbf{z}_{t+\Delta t}|\mathbf{X}_t,\mathbf{X}_{t+\Delta t}),$} & \scalebox{0.8}{\multirow{2}{*}{$\mathcal{D}_{SW}(q_\phi(\Delta \mathbf{z}_t|\mathbf{z}_t),r_\omega(\Delta \mathbf{z}_t|\mathbf{z}_t))$}}\\
 & & & \scalebox{0.8}{$r_\omega(\mathbf{z}_t,\mathbf{z}_{t+\Delta t}))$} & \\
 \multirow{2}{*}{Prior} & \scalebox{0.8}{\multirow{2}{*}{$r_\omega(\mathbf{z})\sim$ Gaussian}} & \scalebox{0.8}{\multirow{2}{*}{$r_\omega(\mathbf{z})\sim$ Uniform}} & \scalebox{0.8}{$r_\omega(\mathbf{z})\sim$ Gaussian} & \scalebox{0.8}{\multirow{2}{*}{$r_\omega(\mathbf{z}_{t+\Delta t}|\mathbf{z}_t)\sim$ Gaussian}}\\
 & & & \scalebox{0.8}{$r_\omega(\mathbf{z}_{t+\Delta t}|\mathbf{z}_t)\sim$ Gaussian} & \\
 \bottomrule
\end{tabular}
\end{table}

\subsection{The inherent dynamics alone can uniquely determine the latent representation.}
To sum up, the workflow so constructed can be summarized through Fig. \ref{fig:alg} and Alg. \ref{algo:rep_learning}. Our algorithm iteratively discretizes the latent space into bins, and resamples the training dataset to focus on poorly sampled regions (see Methods). Then the representation learning is implemented via a two-step optimization process: The first step fixes the prior transition density $r_{\omega}(\Delta\mathbf{z}_t|\mathbf{z}_t)$, and concentrates on learning a better representation by minimizing the representation loss (Eq. \ref{eq:dynamics_rep_obj}); the second step fixes the latent representation and focuses on inferring the transition density $r_{\omega}(\Delta\mathbf{z}_t|\mathbf{z}_t)$ through optimizing the prior loss (Eq. \ref{eq:stochastic_prior_obj}), which in turn provides the information to guide the representation learning process. 

To emphasize the efficacy of our proposed representation learning framework, we leverage recent results on identifiability to demonstrate the model's ability to not only capture the underlying dynamics but also recover the original latent variables, up to an isometry (see Supplementary Note 1 in the Supporting Information for details). This theoretical result offers a profound insight from physics: the intrinsic dynamics alone is sufficient to provide all the information of the system and uniquely determine the latent representation.

In order to provide a quantitative comparison between the proposed DynAE method and other competing algorithms (Table \ref{tab:models}), we measure the isometry-corrected mean square error (MSE) between the true latent representation with the one estimated by the algorithms as done in Ref. \citenum{Hasan2021}. As theoretically proved in Supplementary Note 1 in the Supporting Information, the true latent space and the one obtained by the the proposed DynAE can be equal up to an isometry. Thus, we measure the MSE using the following formula:
\begin{equation} 
\begin{aligned}
\label{eq:MSE_latent}
\mathcal{L}_{z}=\min_{Q,b}\mathbb{E}_{\hat{p}(\mathbf{X})}||Q\phi(\mathbf{X})+b-\mathbf{z}_{true}||_2^2.
\end{aligned}
\end{equation}
Here, the minimum is over all orthogonal matrices $Q \in \mathbb{R}^{d\times d}$ and vectors $b \in \mathbb{R}^{d}$, and $\phi$ is the function learned by the encoder.

\begin{figure}[thb]
    \centering
    \includegraphics[width=0.4\textwidth]{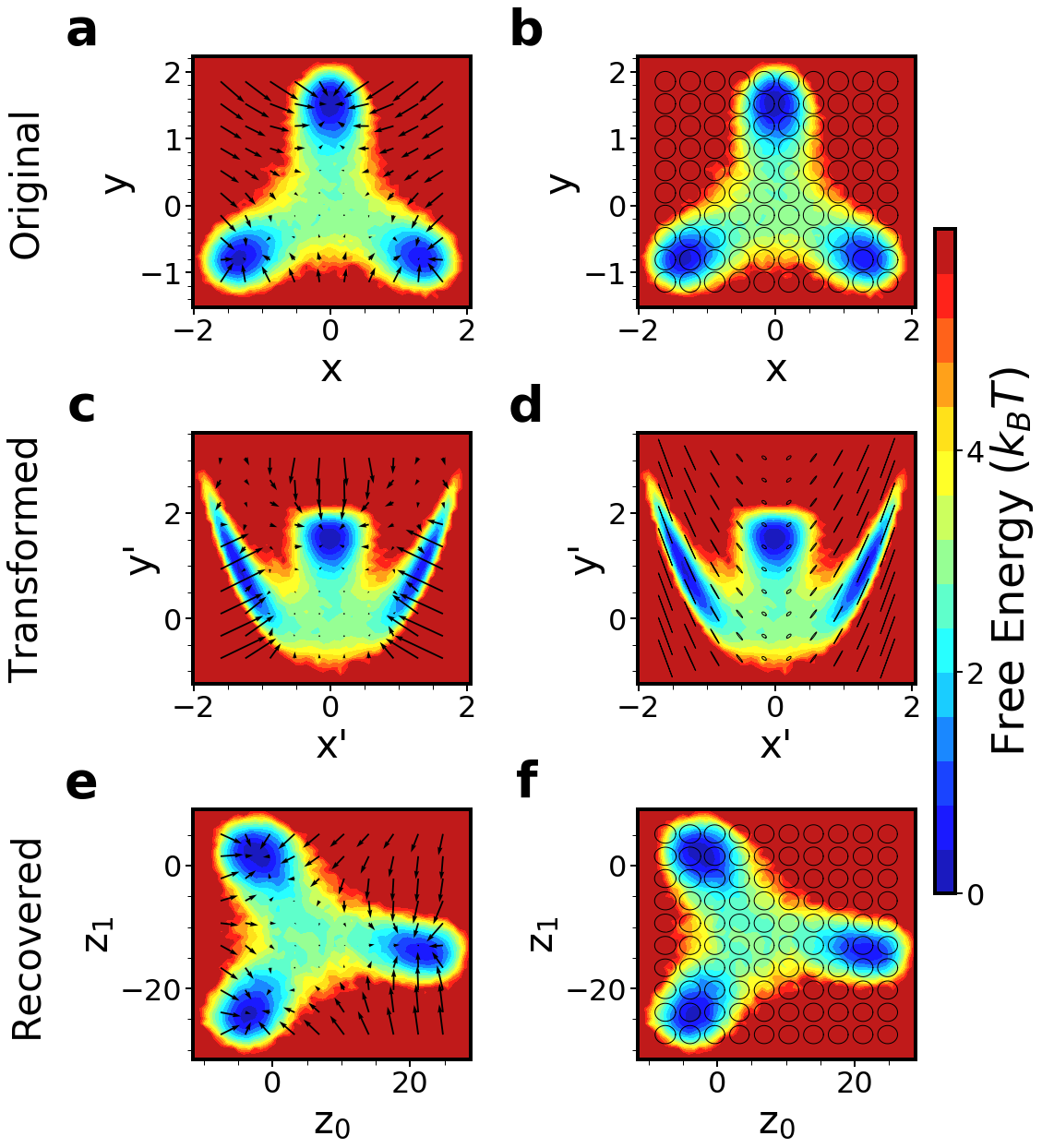}
    \caption{Recovering the underlying dynamics from the transformed three well model system. The original simulated data (\textbf{a,b}), the transformed data (\textbf{c,d}), and the latent representation learned by our algorithm (\textbf{e,f}) are shown in the figure. The black arrows represent the force field $\mathbf{f}=-\nabla F$ (left) while the ellipses represent the diffusion field $\mathbf{M}$ (right). The ellipses in \textbf{d} are highly stretched due to the extremely anisotropic and inhomogeneous diffusion field caused by the nonlinear mapping function.}
    \label{fig:three_state_results}
\end{figure}

\subsection{Recovering the underlying dynamics for three-well model potential.} To illustrate the effectiveness of our approach, we begin by applying it to simulated data generated from a 2D three-well model potential. Prior to inputting this data into our algorithm, we introduce nonlinear mapping functions to simulate the distortions that often occur in real-world observations (see Supplementary Methods in the Supporting Information). The simulated data and its transformation are shown in Fig. \ref{fig:three_state_results}. In the original data space $x$-$y$, the diffusion is constant (Fig. \ref{fig:three_state_results}b). But after the transformation, the diffusion varies dramatically with different $x'$ values as shown in Fig. \ref{fig:three_state_results}d. Feeding this transformed trajectory data into our algorithm, we find our algorithm can successfully recover the original variables up to an invertible linear transformation, wherein the diffusion is once again isotropic and homogeneous. The results also demonstrate the ability of our algorithm to recover the underlying kinetics even in the rarely sampled regions, such as the transition state in the three well model system.

\begin{figure}[b]
    \centering
    \includegraphics[width=0.9\textwidth]{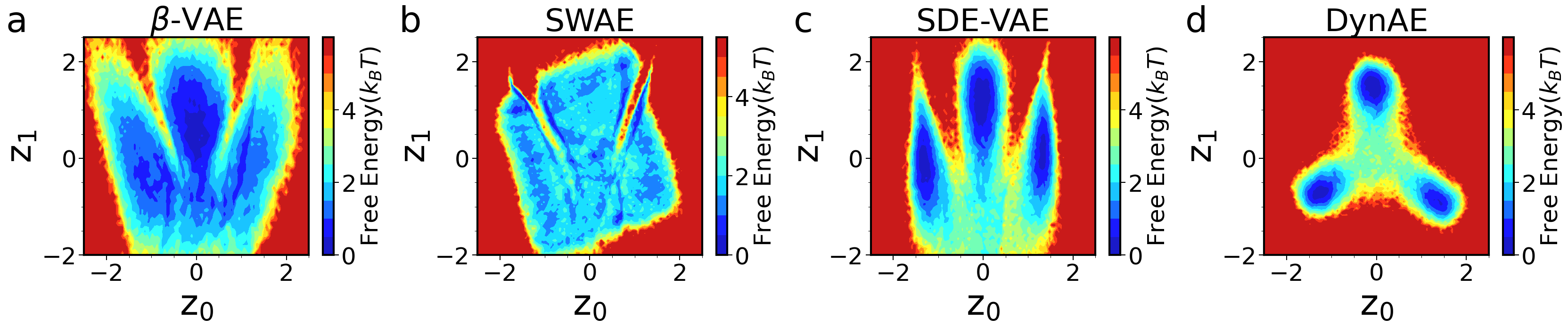}
    \caption{Comparison of the representations learned by different models on the three well model potential. \textbf{a}. $\beta$-VAE. \textbf{b}. SWAE. \textbf{c}. SDE-VAE. \textbf{d}. DynAE. For better comparison, all the representations shown here have been aligned with the ground-truth using the optimum $Q$ from Eq. \ref{eq:MSE_latent}. Only DynAE can successfully recover the ground-truth variables up to isometry.}
    \label{fig:three_state_comparison_results}
\end{figure}

\begin{figure}[t!]
    \centering
    \includegraphics[width=0.4\textwidth]{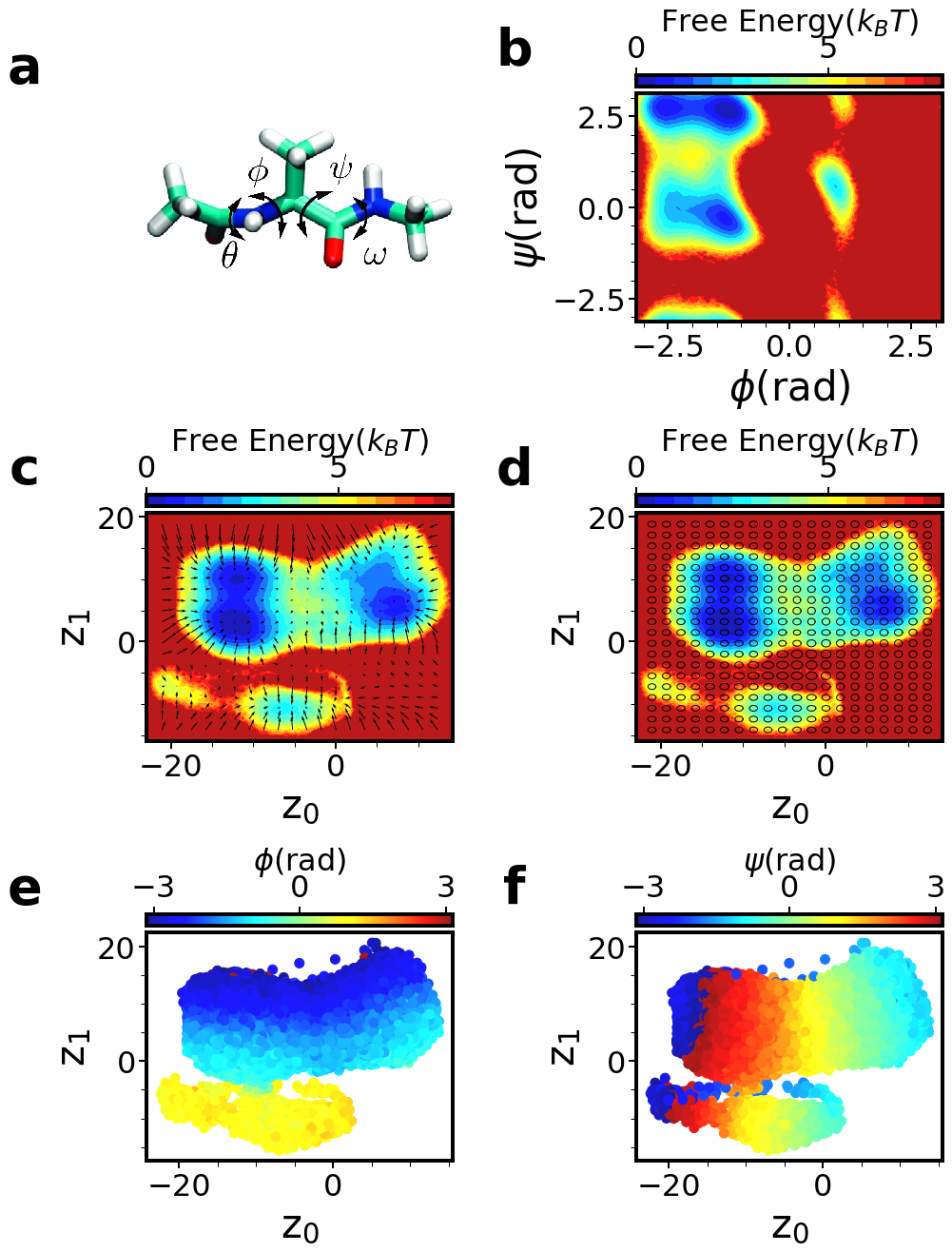}
    \caption{Learning the underlying dynamics from the thirty-dimensional coordinates of alanine dipeptide in water. \textbf{a}. Structure of alanine dipeptide. The main coordinates describing slow transitions are the torsion angles $\phi$ ($C$-$N$-$C_\alpha$-$C$) and $\psi$ ($N$-$C_\alpha$-$C$-$N$), but the neural network input is only the Cartesian coordinates of the heavy atoms. \textbf{b}. Free energy surface of alanine dipeptide in water at 300K along the dihedral angles $\phi$ and $\psi$. \textbf{c, d} show the latent representation learned by our algorithm. The black arrows represent the force field $\mathbf{f}=-\nabla F$ (\textbf{c}) while the ellipses represent the diffusion field $\mathbf{M}$ (\textbf{d}). \textbf{e, f} illustrate the relationship between our learned latent representation and the ground-truth latent factors $\phi$ and $\psi$.}
    \label{fig:aladip_results}
\end{figure}

\begin{figure}[t!]
    \centering
    \includegraphics[width=0.9\textwidth]{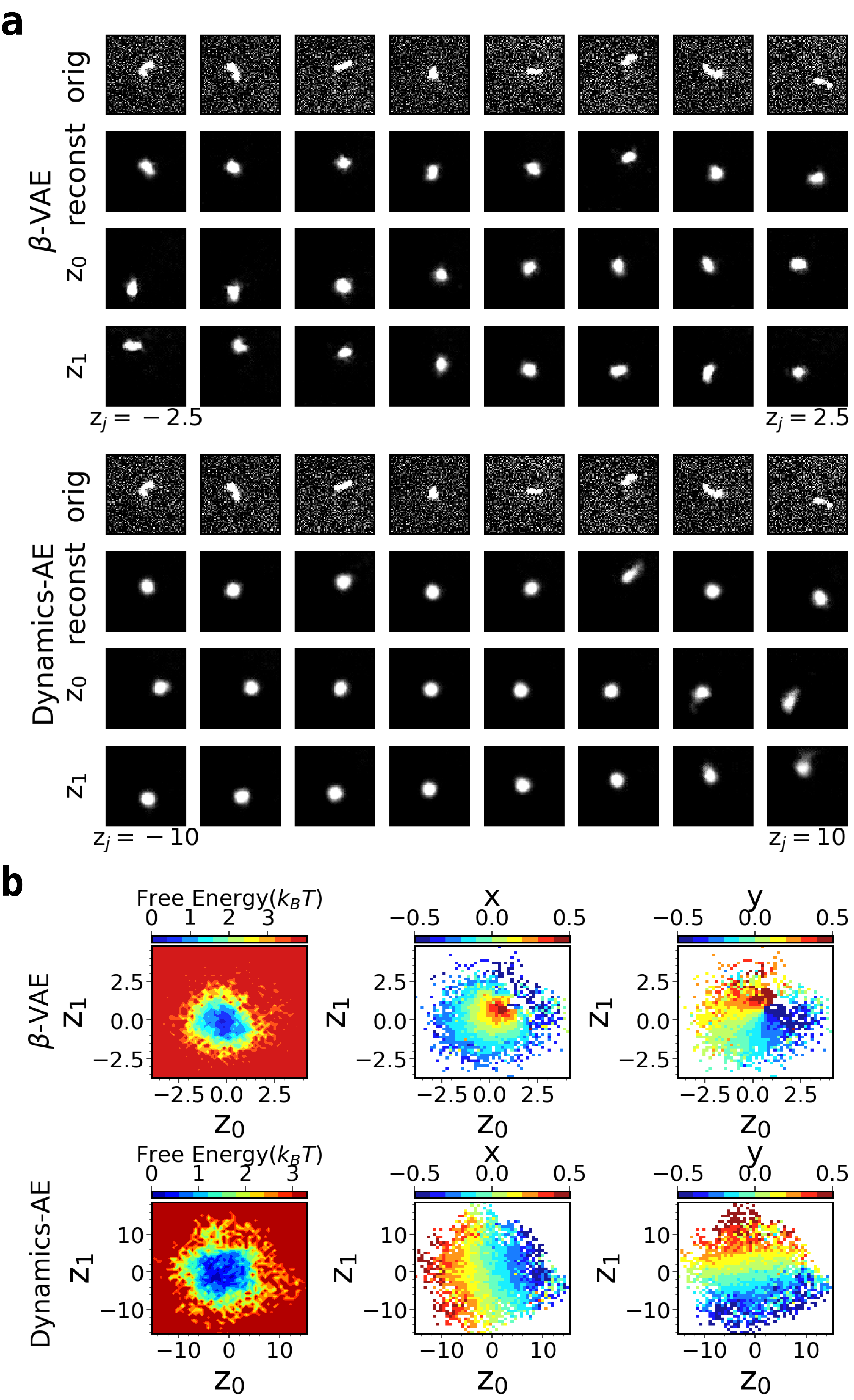}
    \caption{
    Comparison of the results on DNA dataset obtained from $\beta$-VAE (\textbf{a},\textbf{e}), SWAE (\textbf{b},\textbf{f}), SDE-VAE (\textbf{c},\textbf{g}) and DynAE (\textbf{d},\textbf{h}). \textbf{a}-\textbf{d}. Reconstructions of the DNA molecule position. First row: originals. Second row: reconstructions. Remaining two rows: reconstructions of latent traversals across each latent dimension. It can be seen how $z_0$ and $z_1$ for SDE-VAE and DynAE do a much better job than $\beta$-VAE and SWAE at correlation with the underlying movement along $x$ and $y$ directions. \textbf{e}-\textbf{h} The acquired latent representations of the fluorescent DNA molecule and their correlation with the DNA molecule's $x$ and $y$ movements.}
    \label{fig:DNA_results}
\end{figure}

In Figure \ref{fig:three_state_comparison_results}, we present a comparison between the representations acquired by our algorithm and three alternative methods (see Table \ref{tab:models}). $\beta$-VAE and SWAE employ latent space regularization based on predefined probability distributions (Fig. \ref{fig:three_state_comparison_results}a,b): a Gaussian distribution for $\beta$-VAE and a uniform distribution for SWAE. Since the actual ground-truth variables do not follow either a single Gaussian or a uniform distribution, these models undoubtedly fail to learn meaningful representations. By incoporating the dynamics into VAE, SDE-VAE does show an improvement in enhancing the smoothness of the learned latent space. However, it is worth noting that SDE-VAE still struggles to accurately recover the ground-truth variables. This outcome suggests that SDE-VAE struggles when dealing with latent spaces whose distribution significantly diverges from a Gaussian distribution, highlighting the limitations of its regularization term in precisely enforcing the latent dynamics.

\subsection{Identifying physical variables in atomic resolution biomolecule.}
As a slightly more complex example, we further illustrate the power of our algorithm to learn a meaningful representation in the well-studied alanine dipeptide molecule system. We obtain a molecular dynamics trajectory for this system (details available in Supplementary Methods of the Supporting Information). The backbone torsion angles $\phi$ and $\psi$ are known to be the most important reaction coordinates separating different metastable states of alanine dipeptide in water as shown in Fig. \ref{fig:aladip_results}a-b. Without prior knowledge of the dihedral angles, we directly work with a 30-dimensional space comprising the three-dimensional coordinates of the 10 heavy atoms as input. To avoid issue due to trivial rotation and translation of the whole system, we align all the configurations to the first frame of the molecular dynamics trajectory. Fig. \ref{fig:aladip_results}c-d illustrate that our algorithm is still able to recover the underlying dynamics even with a 2D Euclidean latent space. Even more interestingly, a clear correspondence between our learned latent representation and the torsion angles $\phi$ and $\psi$ is found, as shown in Fig. \ref{fig:aladip_results}e-f. This suggests that DynAE can be used to recover physically meaningful variables such as the dihedral angles in alanine dipeptide.

\begin{table}[ht]

\caption{Quantitative comparison of the isometry-corrected MSE $\mathcal{L}_{z}$ from Eq. \ref{eq:MSE_latent} between $\beta$-VAE, SWAE, SDE-VAE and DynAE across different datasets over 5 random seeds. Further experimental setup details can be found in the Supporting Information.}
\label{tab:mse_results}
\centering
 \begin{tabular}{>{\centering\arraybackslash}p{2cm}>{\centering\arraybackslash}p{0.5cm} >{\centering\arraybackslash}p{2.5cm} >{\centering\arraybackslash}p{2cm} >{\centering\arraybackslash}p{3cm} >{\centering\arraybackslash}p{3cm}} 
 \toprule
 \multirow{2}{*}{\textbf{Dataset}} & \multirow{2}{*}{$d_z$} & \multicolumn{4}{c}{$\mathcal{L}_z$}\\
  &  & $\beta$-VAE & SWAE & SDE-VAE & DynAE\\
 \midrule
 \\[-1em]
    DNA & 2 & $1.10\pm0.04$ & $0.92\pm0.13$ & $0.50\pm0.08$ & $0.39\pm0.06$\\
    dSprites-2 & 2 & $0.022\pm0.003$ & $0.37\pm0.18$ & $0.0020\pm0.0001$ & $0.0020\pm0.0001$\\
    dSprites-3 & 3 & $0.730\pm0.022$ & $0.75\pm0.05$ & $0.63\pm0.03$ & $0.22\pm0.09$\\
    shapes3D & 5 & $0.66\pm0.13$ & $0.31\pm0.15$ & $0.32\pm0.08$ & $0.10\pm0.07$\\
 \bottomrule
\end{tabular}
\end{table}

\begin{figure}[t!]
    \centering
    \includegraphics[width=0.8\textwidth]{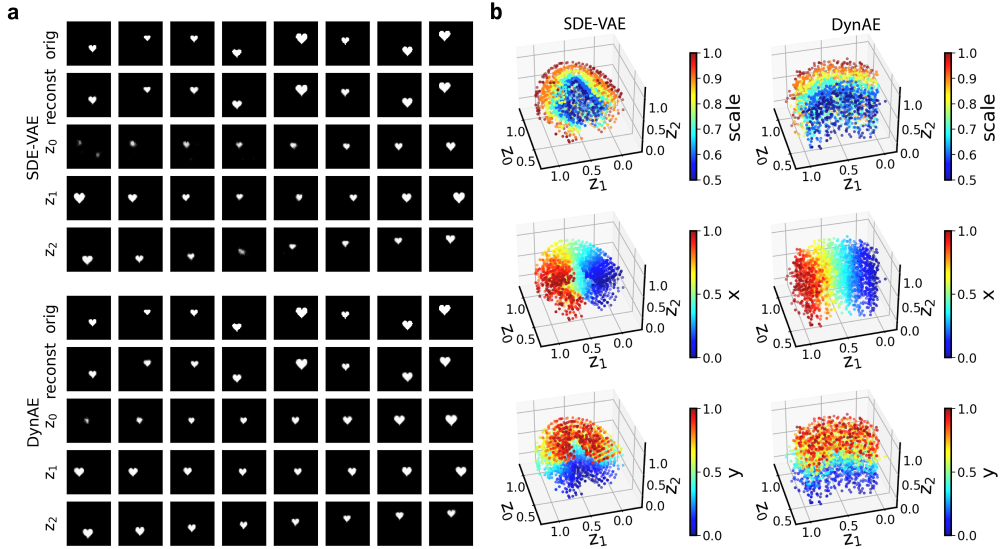}
    \caption{Comparison of the results on dSprites dataset obtained from SDE-VAE and DynAE with three ground-truth factors (Scale, X-Position, Y-Position) of the dataset. \textbf{a}. Reconstructions of dSprites dataset (top: SDE-VAE, bottom: DynAE). First row: originals. Second row: reconstructions. Remaining three rows: reconstructions of latent traversals across each latent dimension. \textbf{b}. Recovering the latent representation from the sequential dSprites images. The left column represents the latent space learned from SDE-VAE. The right column represents the latent representation learned from DynAE.}
    \label{fig:dSprites_3factor_SDEVAE_DynAE_results}
\end{figure}

\subsection{Learning a meaningful representation from fluorescent DNA.} We now apply our algorithm to a real world video dataset consisting of the Brownian motion of DNA molecules in solution as described in the work of Ref. \citenum{Lameh2020}. Following Ref. \citenum{Hasan2021}, we use the same method to obtain the ``ground-truth" latent variables (the $x$ and $y$ coordinates of the center of DNA molecule). The details of the preprocessing protocol are provided in Supplementary Methods of the Supporting Information. As shown in Fig. \ref{fig:DNA_results}a-d, all four algorithms can reconstruct the positions of the molecules with a 2D latent space. However, even with a large $\beta$, $\beta$-VAE and SWAE fail to disentangle the $x$ and $y$ positions in this case. This means a change in a single factor (e.g. $z_1$ in Fig. \ref{fig:DNA_results}a) in the learned representation will lead to a changes in both the $x$ and $y$ positions of the DNA molecule. In comparison, by incorporating dynamics into the learning process, both SDE-VAE and DynAE successfully disentangle the $x$ and $y$ positions in this case. Furthermore, Fig. \ref{fig:DNA_results}e-h clearly illustrate that while $\beta$-VAE and SWAE can only learn a complex nonlinear warping of the ground-truth latent variables, SDE-VAE and DynAE can recover the ground-truth up to an isometry. This highlights the significant advantages of regulating the latent space through its dynamics. However, it's worth noting that SDE-VAE's success in this instance is somewhat expected, given the simplicity of the ground-truth variables and the fortuitous Gaussian distribution they follow.

\subsection{Learning higher dimensional representations.} 
We now extend our algorithm to explore higher-dimensional latent representations using synthesized image datasets with known ground-truth generative factors. We first examine our algorithm in a 2D image dataset (dSprites, see details in the Supporting Information). When only two generative factors (X-Position, Y-Position) of the dSprites dataset are considered, all four algorithms perform admirably. They accurately reconstruct input images and effectively disentangle the two ground-truth factors, as depicted in the Supporting Information (Fig. S1). However, the complexity escalates when we incorporate three generative factors (Scale, X-Position, Y-Position) into the dSprites dataset. This synthesized dataset involves a heart-shaped object exhibiting Brownian motion in the X and Y spatial dimensions while simultaneously changing in size. The results, presented in Fig. \ref{fig:dSprites_3factor_SDEVAE_DynAE_results} and Fig. S2 in the Supporting Information, underscore DynAE's exceptional ability to recover the underlying ground-truth factors, a task that eludes the other algorithms. Notably, in this instance, even though SDE-VAE accounts for dynamics, it struggles to reliably enforce dynamical constraints and uniquely recover the ground-truth variables. Instead, it tends to shape a latent space distribution that closely resembles a Gaussian.

We also test the ability of our algorithm to learn an even higher dimensional latent representation. In this case, five ground-truth factors of the 3D shapes dataset are considered. The results are presented in the Supporting Information (Fig. S3). A quantitative comparison between our method and competing methods across different datasets is summarized in Table \ref{tab:mse_results}. DynAE consistently outperforms or matches the performance of competing methods in terms of recovering the ground-truth latent representations across all the datasets we examined. It's worth highlighting that our evaluation in the synthesized image datasets only considers situations where the ground-truth variables conform to a uniform distribution. However, as exemplified in the three-well model potential, when the ground-truth distribution deviates even further from a Gaussian, such as being multi-modal, we can anticipate an even more pronounced advantage for our algorithm.

\section{Discussion}
In this work, we propose a purely dynamics-constrained representation learning framework inspired by statistical mechanics, enforcing the latent representation to adhere to overdamped Langevin dynamics with a trainable transition density. This approach generalizes recent identifiability results\cite{Hasan2021} to more practical scenarios, and demonstrates its power through a number of numerical examples. In contrast to existing dynamical VAE methods, our algorithm achieves latent space regularization exclusively through intrinsic dynamics, eliminating the need for any prior knowledge about latent space distribution—typically a challenge in complex systems. Additionally, our framework extends the sliced Wasserstein auto-encoder from regularizing the latent variable distribution to regulating the transition density distribution. This subtle adjustment bestows the capability to learn unique representations for a wide spectrum of systems with diverse probability distributions. We believe that this approach, solely centered on constraining latent dynamics, provides a more natural means of latent space regularization. Especially given that the accurate prior probability distribution is usually unavailable, constraining the latent dynamics may arguably be the only generic way to regularize the latent representation.

There are a variety of additional avenues for such a dynamics-based approach. Through both theoretical analysis and numerical experiments, our results further support the previous findings that constraining the latent dynamics is enough to uniquely identify the latent representation up to an isometry \cite{Hasan2021,Hyvarinen2016}. More specifically, we find our algorithm successfully decorrelates the latent representation by assuming a diagonal diffusion matrix. This in fact provides a promising way for dynamics-based disentanglement in representation learning \cite{review_disentanglement}. Besides, our algorithm can also preserve the intrinsic geometry by enforcing a constant homogeneous diffusion field in the latent space. Finally, as only dynamical information matters in the learning process, some more flexible spatial and temporal sampling strategies can be used to make full use of the dataset. In this work, we conduct a preliminary exploration of the sampling scheme by resampling the dataset based on the learned latent representation. Such a sampling strategy allows the algorithm to focus more on poorly sampled regions, greatly improving recovery of the underlying dynamic structure.

This framework provides a completely dynamical viewpoint for representation learning, yielding widely attributed hallmarks of meaningful representations. These include inherent characteristics such as smoothness, disentanglement of explanatory factors, data concentration along low-dimensional manifolds, clusterability, as well as temporal and spatial coherence, as highlighted by previous research\cite{Bengio2013}. In this work, we focused on the overdamped Langevin dynamics and deterministic mappings between observed and latent variables, but the framework should be easily generalized to other dynamic systems, such as the general Langevin dynamics, the deterministic Hamiltonian dynamics and the Schrödinger equation, and many other possible setups including non-deterministic mappings. In other directions, a straightforward extension can be to extend the framework from the autoencoder to the information bottleneck, wherein the latent variables are required only to predict some target $\mathbf{y}$ instead of reconstructing the original input data $\mathbf{X}$ \cite{IB,variational_IB}. The choice of the target $\mathbf{y}$ can be domain-dependent, making the framework highly flexible \cite{PIB,SPIB}. A neural network based likelihood maximization method is used to infer the force field and diffusion field for general purposes, but in principle, other more specific numerical inference method can also be used \cite{Hummer2005, Bullerjahn2020, Sicard2021, Frishman2020}. Another extension would be to generalize our results to non-Euclidean latent space to capture topological properties of certain datasets. In principle, this can be done through some geometric constraints or direct projections to the target manifold \cite{Rey2019}. Finally, one limitation of our approach is that the proposed dynamics constrained regularizer is better suited to continuous deformations of the learned latent space. This regularizer may fail in cases where there are discontinuities in the latent space, which can occur when dealing with highly nonlinear neural networks. To address this issue, we have found that using a smaller learning rate during the initial training phase can help alleviate this problem. Additionally, achieving optimal performance in systems with complicated dynamics may require a sufficiently long lag time ($\Delta t$) to ensure Markovian dynamics in the latent space, along with employing a different, possibly much smaller time step to maintain numerical accuracy in dynamic modeling. Further exploration of these considerations is left for future research.

\section{Methods}

\subsection{Calculating the sliced-Wasserstein distance}
As can be seen from Eq. \ref{eq:obj}, the regularizer $\mathcal{L}_{REG}$ in VAE is the Kullback–Leibler divergence between the posterior distribution $q_\phi(\mathbf{z}|\mathbf{X})$ and the prior distribution $r_\omega(\mathbf{z})$, averaged over the empirical data. However, as discussed in Ref. \citenum{WAE}, this regularizer encourages the encoded distribution for each training sample to become increasingly similar to the prior distribution, fundamentally at odds with the goal of achieving good reconstruction. This can also be shown in the limiting case where $\mathcal{L}_{REG}=\mathcal{D}_{KL}(q_\theta(\mathbf{z}|\mathbf{X}),r_\omega(\mathbf{z}))=0$. This equality occurs if and only if $q_\theta(\mathbf{z}|\mathbf{X})=r_\omega(\mathbf{z})$, suggesting the encoder should forget all information about $\mathbf{X}$. This can lead to difficulties in reconstruction of the input data. Therefore, Ref. \citenum{WAE} proposed Wasserstein Auto-Encoders (WAE) using generative adversarial networks (GAN) to directly match the distribution of the entire encoding space $q_\theta(\mathbf{z}) = \int q_\theta(\mathbf{z}|\mathbf{X})\hat{p}(\mathbf{X}) d\mathbf{X}$ to the prior $r_\omega(\mathbf{z})$. This prevents the latent representations of different data points from collapsing together, thereby promoting better reconstructions while still maintaining simple latent representations. Subsequent research introduced a much simpler solution known as Sliced Wasserstein Auto-Encoders (SWAE) \cite{SWAE}. The trick in SWAE is to project or slice the distribution along multiple randomly chosen directions, and minimize the so-called Wasserstein distance along each of those one-dimensional spaces (which can be done even analytically). Following Ref. \citenum{SWAE}, let $\{\mathbf{z}^n\sim q_\phi(\mathbf{z})\}_{n=1}^N$ and $\{\tilde{\mathbf{z}}^n\sim r_\omega(\tilde{\mathbf{z}})\}_{n=1}^N$ be random samples from the encoded input data and the prior distribution respectively. Assuming $\mathbf{z},\tilde{\mathbf{z}}\in\mathcal{Z}=\mathbb{R}^{d}$, let $\{\theta_l\}_{l=1}^L$ be $L$ randomly sampled directions from a uniform distribution on the $(d-1)$-sphere $\mathbb{S}^{d-1}$. Then the 1D projection of the latent representation $\mathbf{z}$ onto $\theta_l$ is $\mathbf{\theta}_l\cdot \mathbf{z}$. Therefore, the regularizer $\mathcal{L}_{REG}$ in Eq. \ref{eq:obj} can be measured by the sliced-Wasserstein distance:
\begin{equation} 
\begin{aligned}
\label{eq:SW_REG}
\mathcal{L}_{REG}&=\mathcal{D}_{SW}(q_\phi(\mathbf{z}),r_\omega(\mathbf{z})) \\
& = \frac{1}{LN}\sum_{l=1}^L\sum_{n=1}^N||\mathbf{\theta}_l\cdot \tilde{\mathbf{z}}^{i[n]}-\mathbf{\theta}_l\cdot \mathbf{z}^{j[n]}||_2^2,
\end{aligned}
\end{equation}
where $i[n]$ and $j[n]$ are the indices of sorted $\mathbf{\theta}_l\cdot \tilde{\mathbf{z}}^{i[n]}$ and $\mathbf{\theta}_l\cdot \mathbf{z}^{j[n]}$ with respect to $n$, correspondingly. 

\begin{figure}[htb]
\begin{algorithm}[H]
    \caption{Dynamics constrained representation learning}\label{algo:rep_learning}
    \begin{algorithmic}[1]
        \Require{a long unbiased trajectory with the input $\{\mathbf{X}^n\}_{n=1}^N$, latent space dimensionality $d$, number of projections $L$ to approximate sliced Wasserstein distance, minimal distances between cluster centers $d_{\min}$, batch size $B$, $\beta$, $\gamma$}
        \For{epoch $=0,1,2,\dots$}
            \If{epoch $>$ 1}
                \State{Encode the input data $\{\mathbf{z}^n=\phi(\mathbf{X}^n)\}$}
                \State{Discretize the latent space into bins $\{\mathcal{Z}_k\}_{k=1}^K=\text{RegSpaceClustering}(\{\mathbf{z}^n\},d_{\min})$}
                \State{Resample dataset by drawing $N_k\propto\hat{N}_k^{\frac{1}{\gamma}}$ samples from each bin $\{\mathcal{Z}_k\}_{k=1}^K$}
            \EndIf
            \For{$i=0,1,2,\cdots,N/B$}
                \State{Sample a minibatch $\{\mathbf{X}_t^{n_k},\mathbf{X}_{t+1}^{n_k}\}$ of size $B$ from a randomly picked bin}
                \State{Encode the input samples $\{\mathbf{z}_t^{n_k}=\phi(\mathbf{X}_t^{n_k}),\mathbf{z}_{t+1}^{n_k}=\phi(\mathbf{X}_{t+1}^{n_k})\}$}
                \State{Generate prior samples $\{\Delta\tilde{\mathbf{z}}^{n_k}\sim r_\omega(\Delta\mathbf{z}^{n_k}_t|\mathbf{z}^{n_k}_{t})\}$}
                \State{Calculate the objective function $\mathcal{L}_{rep}$}
                \State{Update the neural network parameters $\phi$, $\psi$ through backpropagation} 
                \State{Calculate the objective function $\mathcal{L}_{prior}$}
                \State{Update the neural network parameters $\omega$ through backpropagation}
            \EndFor
        \EndFor
    \end{algorithmic}
\end{algorithm}
\end{figure}

In this work, we want to match the encoded transition density $q_\phi(\Delta\mathbf{z}|\mathbf{z}_t)$ to some specific prior transition density $r_\omega(\Delta\mathbf{z}|\mathbf{z}_t)$. This can be done by minimizing the sliced-Wasserstein distance between these two conditional probability distributions $\mathcal{D}_{SW}(q_\phi(\Delta\mathbf{z}_t|\mathbf{z}_t),r_\omega(\Delta\mathbf{z}_t|\mathbf{z}_t))$ if enough samples from each distribution are available. Samples from the encoded transition density $q_\phi(\Delta\mathbf{z}|\mathbf{z}_t)$ can be obtained, for instance, by performing swarms of short trajectories starting from each point $\mathbf{z}_t$. In many cases, however, these burst trajectories are usually unavailable. Therefore, it would be desirable if this approach could also be applied to a single long trajectory. Here we approximate the sliced-Wasserstein distance $\mathcal{D}_{SW}(q_\phi(\Delta\mathbf{z}_t|\mathbf{z}_t),r_\omega(\Delta\mathbf{z}_t|\mathbf{z}_t))$ by discretizing the latent space $\mathcal{Z}$ into $K$ bins $\{\mathcal{Z}_k\}_{k=1}^K$. In other words, instead of forcing $\Delta\mathbf{z}$ to follow some specific distribution for every point $\mathbf{z}$, we encourage $\Delta\mathbf{z}$ within each bin $\mathcal{Z}_k$ to follow some specific distribution. When the number of bins $K$ is large enough, this should provide a reasonable approximation. The discretization and resampling schemes are discussed in the next subsection. We now can generalize this regularizer loss $\mathcal{L}_{REG}$ to regularize dynamics: 
\begin{equation} 
\begin{aligned}
\label{eq:dynamics_REG_SW}
&\mathcal{L}_{REG}=\mathcal{D}_{SW}(q_\phi(\Delta\mathbf{z}_t|\mathbf{z}_t),r_\omega(\Delta\mathbf{z}_t|\mathbf{z}_t)) \\
& \approx \frac{1}{K}\sum_{k=1}^K\frac{1}{N_k}\mathcal{D}_{SW}(q_\phi(\Delta\mathbf{z}_t|\mathbf{z}_t\in\mathcal{Z}_k),r_\omega(\Delta\mathbf{z}_t|\mathbf{z}_t\in \mathcal{Z}_k))\\
& = \frac{1}{K}\sum_{k=1}^K\frac{1}{LN_k}\sum_{l=1}^L\sum_{n_k=1}^{N_k}||\mathbf{\theta}_l\cdot \Delta\tilde{\mathbf{z}}_t^{i[n_k]}-\mathbf{\theta}_l\cdot \Delta\mathbf{z}_t^{j[n_k]}||_2^2,
\end{aligned}
\end{equation}
where $\Delta\mathbf{z}_t\sim q_\phi(\Delta\mathbf{z}_t|\mathbf{z}_t\in \mathcal{Z}_k),$ $\Delta\tilde{\mathbf{z}}_t\sim r_\omega(\Delta\tilde{\mathbf{z}}_t|\tilde{\mathbf{z}}_t)\in \mathcal{Z}_k$.

Similarly, the reconstruction loss $\mathcal{L}_{REC}$ in Eq. \ref{eq:dynamics_rep_obj} can also be reformulated in terms of bins $\{\mathcal{Z}_k\}$: 
\begin{equation} 
\begin{aligned}
\label{eq:dynamics_REC_K}
\mathcal{L}_{REC}=\frac{1}{K}\sum_{k=1}^K\frac{1}{N_k}\sum_{n_k=1}^{N_k}\bigg[||\psi(\phi(\mathbf{X}_t^{n_k}))-\mathbf{X}_t^{n_k}||_2^2\\
+||\psi(\phi(\mathbf{X}_{t+\Delta t}^{n_k}))-\mathbf{X}_{t+\Delta t}^{n_k}||_2^2\bigg].
\end{aligned}
\end{equation}

\subsection{Discretization and resampling schemes}
For simplicity, regular space clustering with Euclidean distance metric is used to discretize the latent space to obtain uniformly distributed bins $\{\mathcal{Z}_k\}_{k=1}^K$ \cite{Prinz2011}. More advanced distance metrics such as the Mahalanobis distance \cite{Singer2008} and others \cite{Tsai2021} are left for future exploration. The clustering is performed in such a way that cluster centers are at least $d_{\min}$ from each other according to the given metric. Then samples can be assigned to cluster centers through Voronoi partitioning. The final cluster centers obtained are shown in Fig. S4. 

To make the algorithm pay more attention to rarely sampled regions in the latent space, we resample the dataset according to a well-tempered distribution \cite{Barducci2008} by introducing a hyperparameter $\gamma\ge1$:
\begin{equation} 
\begin{aligned}
\label{eq:N_k}
N_k = \frac{\hat{N}_k^{\frac{1}{\gamma}} N}{\sum_{j=1}^K\hat{N}_j^{\frac{1}{\gamma}}},
\end{aligned}
\end{equation}
where $\hat{N}_k$ is the actual number of samples within the bin $\mathcal{Z}_k$ and $N$ is the total number of samples. For $\gamma=1$, Eq. \ref{eq:N_k} reduces to $N_k=\hat{N}_k$ and the original dataset is used for training. For $\gamma\to\infty$, we obtain $N_k=N/K$, which means that we draw samples uniformly from each bin. In practice, we find our results are robust to the choice of $\gamma$, and a moderate value $\gamma=2$ is chosen in this paper.

\subsection{Neural network architecture and training}
The full DynAE algorithm consists of four neural networks: the encoder $\phi$, the decoder $\psi$, and the neural networks learning the diffusion field $\mathbf{M}_\omega$ and the force field $\mathbf{f}_\omega$. For non-image datasets, we use fully connected neural networks with 3 hidden layers and ReLU activations to parameterize the encoder and decoder. Each hidden layer in both the encoder and decoder has 32 nodes (Table S1). For image datasets, we use a convolutional neural network for the encoder and a deconvolutional neural network for the decoder, shown in Table S2. To make a fair comparison, we use the same encoder and decoder architectures for DynAE and all competing methods. In all experiments for DynAE, we parameterize the diffusion and force field using identical architectures: 3-layer fully connected neural networks featuring 32 hidden units and Tanh activation functions (Table S3). It's worth noting that this same neural network is also employed for parameterizing the force field in SDE-VAE. Furthermore, for both DynAE and SWAE experiments, we use 50 projections to calculate the sliced Wasserstein distances. Some other hyper-parameters are also shown in Table S4. Adam optimizers with the same learning rate of 0.001 are used to train the neural networks across all methods.

For the hyper-parameters of DynAE, the minimal distance between cluster centers $d_\text{min}$ should not be too small, ensuring there are enough samples in each bin. The hyper-parameter $\beta$ plays a crucial role in the success of representation learning. From our experiments, we found that the optimal value for $\beta$ should be chosen such that the regularization term, denoted as $\beta_{\mathcal{L}{REG}}$, is on a comparable or even larger scale than the reconstruction error ($\mathcal{L}_{REC}$). If $\beta$ is too small, it may fail to enforce latent dynamics, causing deviations from the desired dynamics. Conversely, an excessively large $\beta$ may prioritize regularization at the expense of faithful data reconstruction, leading to a substantial increase in the reconstruction error. The final reconstruction error is plotted against our performance metric $\mathcal{L}_z$ in Fig. S5 with different $\beta$ over 5 random seeds.

\section{Data availability}
The data for the three-well model potential is available at https://github.com/tiwarylab/DynamicsAE. The fluorescent DNA image dataset is available at https://dataverse.harvard.edu/dataset.xhtml?
persistentId=doi:10.7910/DVN/OJHYZA. The dSprites dataset is available at \\
https://github.com/deepmind/dsprites-dataset. The Shape3D dataset is available at \\
https://github.com/deepmind/3d-shapes. The trajectory data for alanine dipeptide is available from the corresponding author upon reasonable request. 

\section{Code availability}
The python codes using Pytorch will be made available for public use at \\
https://github.com/tiwarylab/DynamicsAE. 

\begin{acknowledgement}

This research was entirely supported by the U.S. Department of Energy, Office of Science, Basic Energy Sciences, CPIMS Program, under Award DE-SC0021009. We also thank Deepthought2, MARCC and XSEDE (projects CHE180007P and CHE180027P) for computational resources used in this work. 

\end{acknowledgement}

\begin{suppinfo}

The Supporting Information is available free of charge at \url{URL_will_be_inserted_by_publisher}.

The Supporting Information provides a theoretical justification for identifiability alongside supplementary results. It also offers technical details of implementation, neural network architecture, dataset preparation, and other relevant information.

\end{suppinfo}

\bibliography{references}

\begin{figure}[h]
    \centering
    \includegraphics[width=9cm]{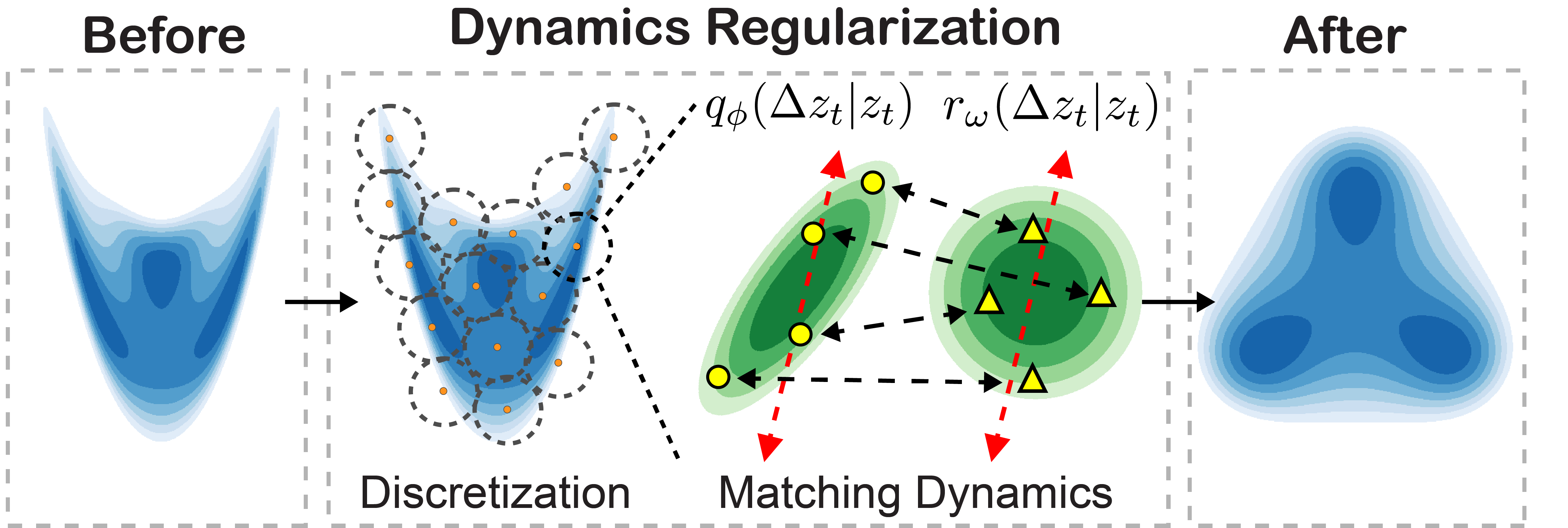}
    \caption*{TOC}
    \label{fig:toc}
\end{figure}

\end{document}